%% file: acl_latex.tex
\documentclass[11pt]{article}

\usepackage[preprint]{acl}

\usepackage{times}
\usepackage{latexsym}

\usepackage[T1]{fontenc}
\usepackage[utf8]{inputenc}

\usepackage{microtype}

\usepackage{inconsolata}

\usepackage{graphicx}

\usepackage{amsmath}
\usepackage{booktabs}
\usepackage{tabularx}
\usepackage{tikz}
\usetikzlibrary{arrows.meta,decorations.pathreplacing,positioning}
\usepackage{multirow}
\usepackage{xcolor}
\usepackage{amssymb}

\newcommand{\tbd}[1]{#1}
\title{What Can't Jev Decide? Probing the Limits of One-Pass Decision Models}

\author{
 \textbf{Wang Yang\textsuperscript{1}},
 \textbf{Xinpeng Li\textsuperscript{1}},
 \textbf{Yanyan Zhang\textsuperscript{1}},
 \textbf{Chuang Ma\textsuperscript{2}},
 \textbf{Shouren Wang\textsuperscript{1}},
 \textbf{Jianzhe Wei\textsuperscript{3}},
\\
 \textbf{Liqin Ye\textsuperscript{3}},
 \textbf{Xiaoran Cai\textsuperscript{4}},
 \textbf{Mohsen Hariri\textsuperscript{1}},
 \textbf{Vipin Chaudhary\textsuperscript{1}},
 \textbf{Xiaotian Han\textsuperscript{1}}
\\
 \textsuperscript{1}Case Western Reserve University,
 \textsuperscript{2}NII LLMC (Japan),
 \\
 \textsuperscript{3}Georgia Institute of Technology,
 \textsuperscript{4}Columbia University
}

\begin{document}
\maketitle
\begin{abstract}
Jev and its open replicas choose among the options a caller supplies with a single forward pass, fast enough to decide inside an agent loop. We ask what such a model cannot decide, and find a simple limit: Jev cannot
make a decision that an LLM makes only by writing out intermediate steps. On a benchmark that varies one factor at a
time, Jev finds and selects records without error up to 20k tokens and 64 options, but once the answer depends on
two or more updates of one user, accuracy falls, to \tbd{53}--\tbd{80\%} at sixteen updates, because similar
records interfere. Given a stale summary of the state, Jev follows it. On
real decision items graded by how many thinking tokens a reference LLM needs, Jev answers \tbd{97.5\%} of the
items that one forward pass of that LLM solves, but only \tbd{74\%} of the others. Jev also does not know its limit:
its confidence flags the errors it makes itself but not those it inherits from a stale input, and neither an
explicit nor a trained hand-off option catches the latter. The open models we evaluate share these limits. Checking the state such models rely on requires a model that reasons.
\end{abstract}

\input{sections/01_intro}

\input{sections/02_models}

\input{sections/03_diagnosis}

\input{sections/04_real}

\input{sections/05_handoff}

\input{sections/06_related}

\input{sections/07_conclusion}

\clearpage
\section*{Limitations}
Our items are isolated decisions: each is posed once, answered once, and scored once. In a deployed agent loop a
decision model is called many times along a trajectory, its errors compound, and they interact with the
surrounding system, with retries, with memory, and with the state summaries that earlier steps wrote, the very
inputs Section~\ref{sec:handoff} shows it does not check. Measuring these effects requires an end-to-end
evaluation of a full agent system, which no single-model benchmark provides. Our results give the per-decision
error rates and the routing signals that such an evaluation would build on.

\bibliography{custom}

\appendix

\input{sections/appendix}

\end{document}

%% file: sections/01_intro.tex
\section{Introduction}
\label{sec:intro}

Jev~\citep{typesafe2026jev} reads a state, a question and a list of options supplied by the caller, and returns a
probability for each option from a single forward pass, in about \tbd{0.2}\,s. It is meant to make the decisions
in agent loops~\citep{yao2023react} while an LLM does the writing. Open replicas followed within
weeks~\citep{kev,jevk5,decider,openjev27b,openjev,litjev,llm2jev}, and public benchmarks report high accuracy and
calibrated probabilities~\citep{jevbench,jevals}. Aggregate accuracy, however, leaves the question that matters to
a system built on Jev unanswered: what can a single forward pass not decide?

The limit we find is simple to state: \emph{Jev cannot make a decision that an LLM makes only by writing out
intermediate steps}. It decides what one record says, but not what several records imply together, nor whether
the state it is given is correct, and it does not know when it has crossed this limit. We show this for Jev 1.13.0
and 23 open configurations of Jev-style models in three steps.

\begin{figure}[t]
  \centering
  \input{figures/fig0_teaser}
  \caption{Where Jev's limit lies. Jev decides what one record says, but not what several records imply together
  or whether a state it is given is correct; an LLM that writes intermediate steps decides all three. Accuracy on
  the controlled benchmark up to 8k tokens: event lookup, final state after 16 updates, and rejection of a stale
  summary (Table~\ref{tab:crossmodel}).}
  \label{fig:teaser}
\end{figure}

\textbf{Locating the limit (Section~\ref{sec:diagnosis}).} On a benchmark that varies one factor at a time, Jev
finds and selects records without error, up to 20k tokens and 64 options. Once the answer depends on two or more
updates of one user, accuracy drops, to \tbd{53}--\tbd{80\%} at sixteen updates, because records of other users
interfere; neither length nor the wording of the question is the cause. Given a precomputed summary of the state,
Jev follows it even when it is stale. An LLM that writes its steps does both tasks: with thinking, Qwen3.8-27B
answers \tbd{96\%} of the multi-update items and rejects \tbd{95\%} of the stale summaries up to 8k tokens, against
\tbd{77\%} and \tbd{74\%} in one pass.

\textbf{The same limit on real tasks (Section~\ref{sec:real}).} We grade 1,131 real decision items by how many
thinking tokens Qwen3.8-27B needs to solve them, and keep the \tbd{738} with reliable labels. Jev's
accuracy falls from \tbd{99\%} on items that need fewer than
250 tokens to \tbd{50\%} on items that need more than 4k; overall it answers \tbd{97.5\%} of the items one forward
pass of Qwen3.8-27B solves, but only \tbd{74\%} of the others. Every open model shows the same gap, and a backbone
trained to reason does not close it.

\textbf{Jev does not know its limit (Section~\ref{sec:handoff}).} Its confidence flags the errors it makes itself
(AUROC \tbd{0.83}) but not those it inherits from a stale input (\tbd{0.53}), and it almost never takes an offered
hand-off option. A hand-off option trained in without new data catches more of its own errors, and is the steadier
routing signal where no labelled data can tune a threshold, but it does not reach the inherited errors either.

Current Jev-style models are therefore selectors over a state computed elsewhere: decisions that need intermediate
steps, including checking that state, belong to a model that reasons. We contribute a controlled benchmark that
locates where one-pass decision models fail, a measure of reasoning demand for real decision items, and an
analysis of whether these models notice and hand off their errors.

%% file: figures/fig0_teaser.tex
% Teaser figure for the introduction: where Jev's limit lies. Drawn in TikZ.
\begin{tikzpicture}[
    font=\scriptsize,
    x=1mm, y=1mm,
    other/.style={draw=black!35, line width=0.45pt},
    target/.style={draw=tgtblue, line width=0.9pt},
    head/.style={font=\scriptsize\sffamily, align=center, inner sep=0},
    val/.style={font=\tiny, inner sep=0.4mm},
  ]
  \definecolor{tgtblue}{HTML}{2A78D6}
  \def\H{13}   % bar height of accuracy 1.0, in mm

  % one column: centre x, heading, Jev value, LLM value, list of target-mark positions in the mini log
  \newcommand{\col}[5]{%
    \node[head] at (#1,31) {#2};
    \draw[black!55, line width=0.35pt] (#1-10,22.2) rectangle (#1+10,25.8);
    \foreach \i in {0,...,19} {\draw[other] (#1-9.5+\i,22.7) -- (#1-9.5+\i,25.3);}
    \foreach \t in {#5} {\draw[target] (#1+\t,22.2) -- (#1+\t,25.8);}
    \fill[tgtblue] (#1-5,4) rectangle (#1-1,4+#3*\H);
    \fill[black!75] (#1+1,4) rectangle (#1+5,4+#4*\H);
    \pgfmathtruncatemacro{\pcta}{round(#3*100)}
    \pgfmathtruncatemacro{\pctb}{round(#4*100)}
    \node[val, above] at (#1-3,4+#3*\H) {\pcta\%};
    \node[val, above] at (#1+3,4+#4*\H) {\pctb\%};
    \draw[black!60, line width=0.35pt] (#1-7,4) -- (#1+7,4);
  }

  \col{11}{one record:\\what does it say?}{1.00}{1.00}{0.5}
  \col{37}{several records:\\what do they imply?}{0.64}{1.00}{-7.5,-2.5,2.5,7.5}
  \col{60}{a given state:\\is it correct?}{0.70}{0.95}{-6.5,-0.5,5.5}
  % the given state is a summary appended to the log
  \draw[black!75, line width=0.5pt] (71,22.2) rectangle (74,25.8);
  \node[font=\tiny, text=black!65, anchor=north east, inner sep=0] at (74,21.6) {summary};

  % the limit
  \draw[black!70, line width=0.5pt, dash pattern=on 1.6pt off 1.2pt] (24,1.5) -- (24,34);
  \node[font=\tiny\itshape, text=black!65, anchor=north west, inner sep=0] at (0.5,2.2) {one forward pass suffices};
  \node[font=\tiny\itshape, text=black!65, anchor=north] at (48.5,2.6) {an LLM writes intermediate steps};

  % legend
  \fill[tgtblue] (0,-5.4) rectangle (2.2,-3.6);
  \node[anchor=west, inner sep=0] at (3,-4.5) {Jev (one pass)};
  \fill[black!75] (25,-5.4) rectangle (27.2,-3.6);
  \node[anchor=west, inner sep=0] at (28,-4.5) {Qwen3.8-27B, thinking};

  % does Jev know?
  \draw[black!30, line width=0.35pt] (0,-8) -- (74,-8);
  \node[anchor=north west, inner sep=0, align=left, text width=74mm] at (0,-9.6)
    {\textbf{Does Jev know?} Its confidence flags its own errors (AUROC \tbd{0.83}) but not errors that come
     from a stale input (\tbd{0.53}).};
\end{tikzpicture}

%% file: sections/02_models.tex
\section{Jev and Jev-Style Decision Models}
\label{sec:models}

\paragraph{Decision models.}
A decision model reads a state $s$, a question $q$ and a set of options $\mathcal{O}$ supplied by the caller, and
returns a probability distribution $p(\cdot \mid s,q,\mathcal{O})$ over $\mathcal{O}$ from a single forward pass,
without generating text. Its answer is $\hat{o}=\arg\max_{o\in\mathcal{O}} p(o)$ and its confidence is
$p_{\max}=p(\hat{o})$.

\paragraph{Jev.}
We evaluate Jev 1.13.0~\citep{typesafe2026jev} through its API, which accepts up to 32k tokens; its weights and
training data are not public.

\paragraph{Jev-style models.}
To tell whether a failure belongs to Jev or to the model class, we also evaluate 23 open configurations that serve
the same interface (listed in Table~\ref{tab:serving}, Appendix~\ref{app:models}): 18 from nine released projects and five plain Qwen models read through
their option-letter logits. Most fine-tune a Qwen backbone~\citep{qwen3,qwen35,qwen36,qwen38} on decision
data~\citep{kev,jevk5,decider,openjev27b,openjev}; the others wrap an unmodified LLM~\citep{litjev,llm2jev,openjevsglang},
read the option-letter logits of the base LLM~\citep{jqv,semif}, or score options contrastively~\citep{clm}.

\paragraph{Protocol.}
Every model sees the same state, question, options and option order. Inputs beyond a model's context limit are
recorded as unsupported rather than truncated, and cross-model comparisons use only the items every compared model
supports.

%% file: sections/03_diagnosis.tex
\section{Where Jev Breaks: A Controlled Diagnosis}
\label{sec:diagnosis}

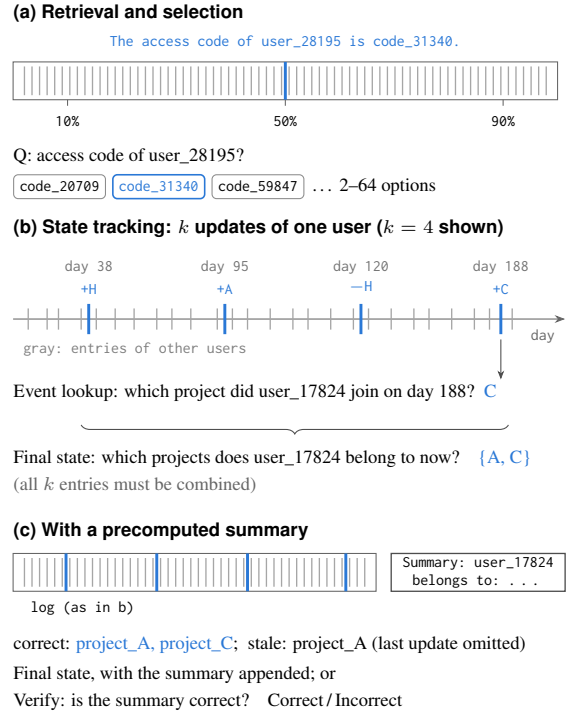
\begin{figure}[t]
  \centering
  \input{figures/fig1_tasks}
  \caption{The three task families. Thin marks are log entries or records; blue marks belong to the target. Options
  are drawn from the context, so every wrong option also occurs in it.}
  \label{fig:bench}
\end{figure}

\subsection{A benchmark that varies one factor at a time}

\paragraph{Design.}
Each item consists of a context, a question and 2--64 options, and its answer is computed by a program rather than
labelled by a model (Figure~\ref{fig:bench}). Contexts range from 1k to 20k tokens. For each scenario, shorter
contexts are prefixes of longer ones and the target records keep their relative position, so neighbouring cells
differ only in the factor under test. Every wrong option also appears in the context, so no option can be ruled out
without reading it. Each cell contains 10 scenarios, each with 3 option orders.

\paragraph{Tasks.}
\emph{Retrieval} asks for the access code of one user, hidden among other users' records or among the essay
paragraphs of the needle-in-a-haystack test~\citep{kamradt2023needle},
at 10\%, 50\% or 90\% depth. \emph{Selection} asks the same question with 2--64 candidate codes, all of which occur in
the context. \emph{State tracking} gives a day-ordered log in which the target user joins and leaves projects $k$
times ($k\in\{1,2,4,8,16\}$), interleaved with other users' logs. Each log is asked two questions: \emph{event lookup}
asks what happened on one day, and \emph{final state} asks which projects the user belongs to after the last entry.
The two questions share the context and differ only in whether the $k$ updates must be combined.

\subsection{Length and the number of options are not the bottleneck}

We first test whether finding and selecting a record degrade with length, depth or the number of options. They do
not: Jev answers all 105 retrieval, selection and event-lookup cells correctly, up to 20k tokens and 64 options, with
the target at any depth and in either haystack, and puts a probability close to 1 on the correct option. Long
contexts, buried targets and many options, the usual explanations for failures of long-context
models~\citep{kamradt2023needle,liu2024lost,hsieh2024ruler}, therefore do not explain the failures below.

\subsection{State tracking fails from two updates on}

\begin{figure*}[t]
  \centering
  \includegraphics[width=\textwidth]{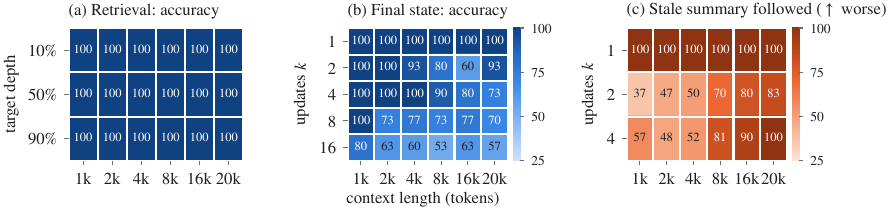}
  \caption{Jev 1.13.0 on the controlled benchmark. (a)~Retrieval is perfect at every depth and length.
  (b)~Final state fails from two updates on, although event lookup on the same logs is perfect.
  (c)~With a stale summary appended, the share of answers that repeat the stale state; only items whose options
  include the stale state, which is rare for $k\geq 8$. At $k=1$ the stale state is the empty set.
  Blue panels show accuracy (higher is better); the orange panel shows a failure rate (higher is worse).
  Each cell: 10 scenarios $\times$ 3 option orders. All panels in \%; colour scales start at 25\%, chance with
  four options.}
  \label{fig:heatmap}
\end{figure*}

We next ask for the user's current state, which depends on all $k$ updates in the same log. Accuracy drops as soon as
$k=2$ and falls to \tbd{53}--\tbd{80\%} at $k=16$ (Figure~\ref{fig:heatmap}b), although Jev answers every event
lookup on these logs correctly. It finds each update when asked about it, but does not use all of them when the
answer depends on all of them.

The errors show which updates are lost: the later ones. Of \tbd{145} wrong answers, \tbd{134} are an earlier state of
the same user and \tbd{137} contain fewer projects than the correct set; only \tbd{3} confuse users. The probability
on the correct option falls before the accuracy does: it averages \tbd{0.93} in the multi-update cells that are
still answered perfectly, and about \tbd{0.5} at $k=16$.

\subsection{Prompting does not help; similar records are the cause}

We first ask whether a better question fixes the failure. It does not (Table~\ref{tab:prompt}, rows P1--P3): asking
for step-by-step thinking (P1), spelling out the procedure of going through the user's entries in order (P2), or
naming the failure observed above (P3) leaves Jev between \tbd{54\%} and \tbd{64\%} at $k=16$, against \tbd{61\%}
for the original item.

We then change the log, keeping the question and options, to find which of its properties causes the failure (rows
R1--R4). With only the target user's entries left (R1), Jev answers every item correctly, so it can combine 16
updates. Moving these entries, in order, into one block at the end of the log
while keeping every other entry (R3) leaves the length and the other users' records unchanged, and accuracy at $k=16$
stays at \tbd{64\%}. Replacing the other users' entries of R3 by essay text of the same length (R4) raises it to
\tbd{99\%}. The failure is caused neither by scattered entries nor by length, but by the similar records around them, a sharper
form of the distraction by irrelevant context known from LLMs~\citep{shi2023large}.
Position does not explain it either: listing the log newest first (R2) lowers accuracy at $k\geq2$ from \tbd{80\%}
to \tbd{72\%}.

This interference differs from the one reported for LLMs, where earlier values of the same key displace the latest
one~\citep{wang2025unable}: R1 contains all of the target's earlier updates and is answered perfectly. What Jev
fails at is keeping one user's updates apart from other users' records. Kev-27B behaves alike, except that R3
already suffices; Kev-4B reaches only \tbd{67\%} at $k=16$ even in R1 and R4, so at this scale the
combination itself is limited as well.

\begin{table}[t]
  \centering
  \small
  \setlength{\tabcolsep}{2.6pt}
  \begin{tabular}{@{}l ccc ccc@{}}
    \toprule
    & \multicolumn{3}{c}{$k\geq2$} & \multicolumn{3}{c}{$k=16$} \\
    \cmidrule(lr){2-4}\cmidrule(l){5-7}
    \textbf{Variant} & Jev & K-27B & K-4B & Jev & K-27B & K-4B \\
    \midrule
    P0 original          & 80 & 90 & 66 & 61 & 86 & 41 \\
    \midrule
    P1 step by step      & 81 & 90 & 67 & 61 & 88 & 39 \\
    P2 procedure         & 79 & 92 & 65 & 54 & 87 & 38 \\
    P3 name failure      & 78 & 88 & 62 & 64 & 86 & 38 \\
    \midrule
    R1 target only       & 100 & 100 & 85 & 100 & 100 & 67 \\
    R2 newest first      & 72 & 88 & 20 & 61 & 80 & 4 \\
    R3 target at end     & 89 & 97 & 61 & 64 & 100 & 29 \\
    R4 R3 + essays       & 100 & 99 & 78 & 99 & 99 & 67 \\
    \bottomrule
  \end{tabular}
  \caption{Final-state accuracy in \% when the question (P1--P3) or the log (R1--R4) is changed. R3 and R4 have the same
  length and target positions as P0; R4 replaces the other users' entries by essay text. Jev: 1k--20k tokens;
  Kev-27B (K-27B) and Kev-4B (K-4B): up to 4k. Kev-9B is in Appendix~\ref{app:bench}.}
  \label{tab:prompt}
\end{table}

In practice the target's history cannot be pulled out of a real log by hand; the usual way to spare the model the
search is to state the result in advance, which we test next.

\subsection{A state summary is used, trusted, and not checked}

We next test the obvious remedy: compute the state in advance and append it to the log as a one-line summary. Jev
uses the summary, but also trusts it. With a \emph{correct} summary, it answers the final-state question correctly
in all 30 cells. With a \emph{stale} summary,
which omits the last update, Jev picks the stale state in \tbd{61--100\%} of the items that offer it as an option, even
though the log contradicts it, and more often in longer logs (\tbd{68\%} at 1k, \tbd{93\%} at 20k;
Figure~\ref{fig:heatmap}c).

Asked directly, Jev cannot tell a correct summary from a stale one. It accepts \tbd{91\%} of correct
summaries but rejects only \tbd{62\%} of stale ones over all context lengths (\tbd{92\%} and \tbd{70\%} up to 8k,
Table~\ref{tab:crossmodel}), and almost none (\tbd{0--7\%}) when the summary omits one of two
updates in a 16k--20k log. The remedy is therefore only as good as the summary.

\subsection{The failures are shared by the model class}

We finally ask whether these failures are specific to Jev. They are not: on contexts up to 8k tokens, every open
model in Table~\ref{tab:crossmodel} shows the same pattern: retrieval and selection are close to perfect, final state drops once updates must be
combined, and a stale summary is followed. No decision model verifies summaries reliably: averaged over correct and
stale summaries, none exceeds \tbd{85\%} (chance 50\%), and the models that reject the most stale summaries reject
most correct ones as well (Qwen3.5-4B-Base rejects \tbd{98\%} of stale and accepts \tbd{2\%} of correct ones).
Neither scale (0.8B to 35B), training on questions written by a reasoning teacher (JevK5), nor reading the logits of
an untuned LLM changes this pattern. Two models in Appendix~\ref{app:bench} fail earlier, at retrieval itself:
Kev-0.8B and the contrastive CLM-8B.

\begin{table*}[t]
  \centering
  \small
  \begin{tabular}{@{}l ccc ccc cc c@{}}
    \toprule
    & \multicolumn{3}{c}{\textbf{Find and select}} & \multicolumn{3}{c}{\textbf{Final state}} & \multicolumn{2}{c}{\textbf{With summary}} & \textbf{Verify} \\
    \cmidrule(lr){2-4}\cmidrule(lr){5-7}\cmidrule(lr){8-9}\cmidrule(l){10-10}
    \textbf{Model} & Retr. & Select & Event & $k{=}1$ & $k{\geq}2$ & $k{=}16$ & correct & stale & reject stale \\
    \midrule
    Jev 1.13.0          & 100 & 100 & 100 & 100 & 84 & 64 & 100 & 58 & 70 \\
    \midrule
    Open-Jev-27B        & 100 & 100 & 100 & 100 & 89 & 78 & 100 & 72 & 74 \\
    OpenJev-27B         & 100 & 100 & 100 & 100 & 95 & 93 & 100 & 66 & 56 \\
    Kev-27B$^\dagger$   & 100 & 100 & 100 & 100 & 91 & 88 & 100 & 60 & 62 \\
    Kev-4B$^\dagger$    & 100 & 100 & 100 & 98 & 66 & 41 & 100 & 53 & 38 \\
    JevK5-4B            & 100 & 100 & 99 & 98 & 53 & 56 & 100 & 47 & 52 \\
    Decider-35B-A3B     & 100 & 100 & 99 & 83 & 65 & 60 & 100 & 39 & 31 \\
    Decider-4B          & 100 & 100 & 98 & 100 & 67 & 80 & 100 & 39 & 0 \\
    LitJev-27B          & 100 & 100 & 100 & 98 & 79 & 76 & 100 & 41 & 7 \\
    llm2jev-27B         & 100 & 100 & 100 & 100 & 86 & 81 & 100 & 43 & 73 \\
    Qwen3.8-27B, letter logits & 100 & 100 & 100 & 100 & 77 & 77 & 100 & 42 & 74 \\
    \midrule
    Qwen3.8-27B, thinking$^\ddagger$ & -- & -- & 100 & 100 & 96 & 100 & 100 & 83 & 95 \\
    \bottomrule
  \end{tabular}
  \caption{Accuracy in \% of Jev and representative open models on contexts up to 8k tokens. ``Select'' pools 2--64
  options. ``With summary'' is final-state accuracy when a correct or a stale summary is appended; ``reject stale''
  is the share of stale summaries judged incorrect (chance 50\%). $^\dagger$Kev-4B and Kev-27B accept at most 4k
  tokens and are evaluated up to 4k. $^\ddagger$Not a decision model: the same LLM generates its reasoning before
  answering; shown for reference and run on the state-tracking items only. All models are in
  Appendix~\ref{app:bench}.}
  \label{tab:crossmodel}
\end{table*}

Jev thus decides what one record says, but not what several records imply together once similar records surround
them, nor whether a state it is given is correct. Both are decisions that an LLM makes by writing intermediate
steps: with thinking, Qwen3.8-27B answers \tbd{96\%} of the multi-update items and rejects \tbd{95\%} of the stale
summaries (Table~\ref{tab:crossmodel}). Section~\ref{sec:real} asks whether the same limit appears on real
decision tasks.

%% file: figures/fig1_tasks.tex
% Figure 1: the three task families. Drawn in TikZ; included from sections/03_diagnosis.tex.
\begin{tikzpicture}[
    font=\scriptsize,
    x=1mm, y=1mm,
    other/.style={draw=black!35, line width=0.5pt},
    target/.style={draw=tgtblue, line width=1.1pt},
    lbl/.style={font=\scriptsize\sffamily\bfseries, anchor=west, inner sep=0},
    txt/.style={anchor=west, inner sep=0, align=left},
    tt/.style={font=\tiny\ttfamily},
    opt/.style={draw=black!45, rounded corners=0.6mm, inner sep=0.8mm, font=\tiny\ttfamily},
    ok/.style={opt, draw=tgtblue, line width=0.6pt, text=tgtblue},
  ]
  \definecolor{tgtblue}{HTML}{2A78D6}

  % ---------- (a) retrieval / selection ----------
  \node[lbl] at (0,64.5) {(a) Retrieval and selection};
  \draw[black!60, line width=0.4pt] (0,53) rectangle (72,58);
  \foreach \i in {1,...,70} {\draw[other] (\i+0.5,53.6) -- (\i+0.5,57.4);}
  \draw[target] (36,53) -- (36,58);
  \foreach \p/\t in {7.2/10\%, 36/50\%, 64.8/90\%} {\draw[black!60] (\p,53) -- (\p,52.2); \node[tt, below] at (\p,52.2) {\t};}
  \node[tt, text=tgtblue, anchor=south] at (36,58.6) {The access code of user\_28195 is code\_31340.};
  \node[txt] at (0,45.5) {Q: access code of user\_28195?};
  \node[opt, anchor=west] (a1) at (0,41.5) {code\_20709};
  \node[ok, anchor=west, right=0.8mm of a1] (a2) {code\_31340};
  \node[opt, anchor=west, right=0.8mm of a2] (a3) {code\_59847};
  \node[txt, right=1mm of a3] {\dots\ 2--64 options};

  % ---------- (b) state tracking ----------
  \node[lbl] at (0,36) {(b) State tracking: $k$ updates of one user ($k=4$ shown)};
  \draw[black!60, line width=0.4pt, -{Stealth[length=1.4mm]}] (0,24) -- (73,24) node[tt, below left] {day};
  \foreach \x in {2,4.5,6,9,11,13.5,16,19,21,23.5,27,29,31,34,37,39,42,45,47,50,53,55,58,61,63,66} {\draw[other] (\x,22.6) -- (\x,25.4);}
  \foreach \x/\l in {10/{+H}, 28/{+A}, 46/{$-$H}, 64.5/{+C}} {
    \draw[target] (\x,22) -- (\x,26);
    \node[tt, text=tgtblue, above] at (\x,26) {\l};
  }
  \foreach \x/\d in {10/38, 28/95, 46/120, 64.5/188} {\node[tt, text=black!55, above] at (\x,28.6) {day \d};}
  \node[tt, text=black!50, anchor=west] at (0,20) {gray: entries of other users};
  % event lookup reads one entry
  \draw[black!70, -{Stealth[length=1.2mm]}] (64.5,21.3) -- (64.5,16);
  \node[txt] at (0,14.2) {Event lookup: which project did user\_17824 join on day 188?\enspace \textcolor{tgtblue}{C}};
  % final state combines all of them
  \draw[decorate, decoration={brace, mirror, amplitude=1.2mm}, black!70] (9,10) -- (65.5,10);
  \node[txt] at (0,5.5) {Final state: which projects does user\_17824 belong to now?\quad \textcolor{tgtblue}{\{A, C\}}};
  \node[txt, text=black!60] at (0,2) {(all $k$ entries must be combined)};

  % ---------- (c) summary ----------
  \node[lbl] at (0,-4) {(c) With a precomputed summary};
  \draw[black!60, line width=0.4pt] (0,-12) rectangle (48,-7);
  \foreach \i in {1,...,46} {\draw[other] (\i+0.5,-11.4) -- (\i+0.5,-7.6);}
  \foreach \x in {7,19,31,44} {\draw[target] (\x,-12) -- (\x,-7);}
  \node[tt, anchor=west] at (1,-14.2) {log (as in b)};
  \draw[black!70, line width=0.5pt] (50,-12) rectangle (73,-7);
  \node[tt, align=center] at (61.5,-9.5) {Summary: user\_17824\\belongs to: \dots};
  \node[txt] at (0,-19) {correct: \textcolor{tgtblue}{project\_A, project\_C};\enspace stale: project\_A (last update omitted)};
  \node[txt] at (0,-23) {Final state, with the summary appended; or};
  \node[txt] at (0,-26.6) {Verify: is the summary correct?\quad Correct\,/\,Incorrect};
\end{tikzpicture}

%% file: sections/04_real.tex
\section{Where Jev Breaks on Real Tasks}
\label{sec:real}

Real tasks offer no factor to vary, so we measure how much reasoning each item demands and relate Jev's accuracy
to it.

\subsection{Items and reasoning demand}

\paragraph{Items.}
We test on 1,131 real decision items with a fixed option list: the public JevBench tiers (easy, original and hard;
231 items;~\citealp{jevbench}) and 300 items each from PubMedQA~\citep{jin2019pubmedqa}, Banking77
\citep{casanueva2020banking77} and HelpSteer2~\citep{wang2024helpsteer2} as packaged by Jevals~\citep{jevals}.
Questions are yes/no, a choice among 2--77 labels, or a score level.

\paragraph{Reasoning demand.}
To measure how much reasoning an item needs, we run Qwen3.8-27B on it in three ways: reading the option probabilities after one forward pass without
generating (\emph{direct readout}), answering after generation without thinking, and answering after thinking, the
latter two with 10 samples each. An item is \emph{one-pass solvable} if the direct readout is correct, and
\emph{needs reasoning} otherwise. Its \emph{demand} is the median number of thinking tokens in the correct thinking
samples. Qwen3.8-27B is only a reference for how much reasoning an item requires, not a model of Jev, whose backbone is
not public.

\begin{figure}[t]
  \centering
  \input{figures/fig3_examples}
  \caption{Two real-task items, with the state abbreviated. (a)~An item that needs a short calculation, which Jev
  gets wrong in one pass. (b)~One of the \tbd{93} items excluded because their gold label is disputed: both models
  give the same other label.}
  \label{fig:realex}
\end{figure}
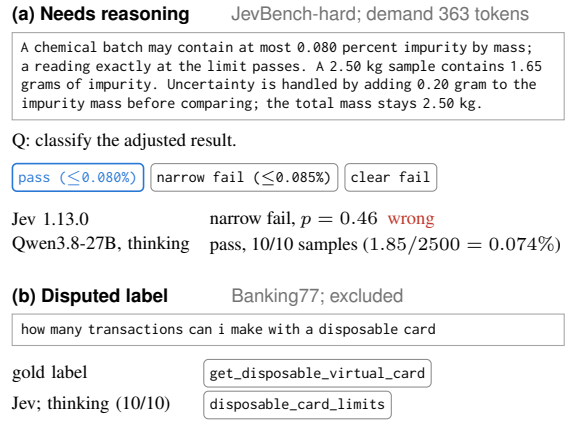

\paragraph{Excluded items.}
We set aside items whose labels cannot separate the models. HelpSteer2 asks for one of five helpfulness
levels, and no model clearly beats always choosing the most frequent level (\tbd{42\%}): the decision models reach
\tbd{31}--\tbd{49\%} and Qwen3.8-27B with thinking \tbd{33\%}, so we report it only in Appendix~\ref{app:bench}.
We further exclude the \tbd{93} items that Qwen3.8-27B with thinking answers correctly in at most 5 of 10 samples:
\tbd{65} from Banking77, \tbd{25} from PubMedQA and \tbd{3} from JevBench-hard. Their gold labels are disputed rather than hard: Jev
answers \tbd{76} of them incorrectly, and in \tbd{68} of these it picks the same wrong answer as the majority of the
thinking samples (example in Figure~\ref{fig:realex}b). The remaining \tbd{738} items comprise \tbd{646} one-pass solvable items and \tbd{92}
that need reasoning. Results on all items are in Appendix~\ref{app:bench}.

\subsection{Jev loses the items that need reasoning}

\begin{figure}[t]
  \centering
  \includegraphics[width=\columnwidth]{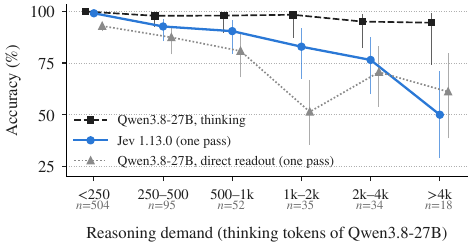}
  \caption{Accuracy on the \tbd{738} real-task items against the number of thinking tokens Qwen3.8-27B needs to
  answer them. Jev and the direct readout answer in one forward pass; the thinking model is averaged over 10
  samples. Items are selected by the thinking model (Section~\ref{sec:real}), so its curve is an upper reference,
  not a competitor. Error bars: 95\% Wilson intervals.}
  \label{fig:demand}
\end{figure}

Jev loses accuracy where an item needs reasoning, by every measure we use. Its accuracy falls with the demand of
the item (Figure~\ref{fig:demand}), from \tbd{99\%} on items that need
fewer than 250 thinking tokens to \tbd{50\%} on items that need more than 4k, while Qwen3.8-27B with thinking stays
above \tbd{94\%}. Jev answers \tbd{97.5\%} of the one-pass solvable items but only \tbd{74\%} of those that need
reasoning. The gap sits in one source (Table~\ref{tab:real}): on JevBench-hard, whose items apply policies, dates
and multi-step rules, Jev reaches \tbd{73\%} against \tbd{99\%} for the thinking model, and comes within about one point of it on the
other four sources. Of Jev's \tbd{40} errors, \tbd{24} are on items that need reasoning; of the other \tbd{16}, \tbd{8} are
JevBench-hard items that the direct readout happens to answer correctly.

\subsection{The model class shares it}

We next ask whether the open models show the same gap. They do (Table~\ref{tab:real}): every model is close to its
best on the one-pass solvable items and makes most of its errors on the items that need reasoning: the 27B models answer
\tbd{62}--\tbd{70\%} of them, and the 4B and 9B models \tbd{53}--\tbd{60\%}. For every model the gap
between the two kinds of items lies between \tbd{21} and \tbd{38} points, and Jev has one of the smallest.

\begin{table*}[t]
  \centering
  \small
  \begin{tabular}{@{}l cc ccccc c@{}}
    \toprule
    & \multicolumn{2}{c}{\textbf{By demand}} & \multicolumn{5}{c}{\textbf{By source}} & \\
    \cmidrule(lr){2-3}\cmidrule(lr){4-8}
    \textbf{Model} & One pass & Reasoning & J-easy & J-original & J-hard & PubMedQA & Banking77 & \textbf{All} \\
    & \tbd{646} & \tbd{92} & \tbd{48} & \tbd{72} & \tbd{108} & \tbd{275} & \tbd{235} & \tbd{738} \\
    \midrule
    Jev 1.13.0          & 97.5 & 73.9 & 100.0 & 98.6 & 73.1 & 98.5 & 97.4 & 94.6 \\
    \midrule
    Kev-27B             & 97.4 & 69.6 & 100.0 & 100.0 & 72.2 & 96.7 & 97.4 & 93.9 \\
    llm2jev-27B         & 95.4 & 68.5 & 100.0 & 95.8 & 74.1 & 94.5 & 94.5 & 92.0 \\
    LitJev-27B          & 95.4 & 63.0 & 100.0 & 98.6 & 75.9 & 93.5 & 91.9 & 91.3 \\
    Open-Jev-27B        & 83.4 & 62.0 & 100.0 & 95.8 & 74.1 & 75.6 & 81.3 & 80.8 \\
    JevK5-9B            & 94.6 & 59.8 & 100.0 & 95.8 & 77.8 & 94.2 & 87.7 & 90.2 \\
    Kev-4B              & 91.2 & 53.3 & 100.0 & 93.1 & 52.8 & 88.4 & 94.9 & 86.4 \\
    \midrule
    \multicolumn{9}{@{}l}{\emph{Reference: Qwen3.8-27B}} \\
    Direct readout      & 100.0 & 0.0 & 97.9 & 94.4 & 68.5 & 96.4 & 81.7 & 87.5 \\
    Thinking            & 99.3 & 97.3 & 100.0 & 99.7 & 99.2 & 98.9 & 98.6 & 99.0 \\
    \bottomrule
  \end{tabular}
  \caption{Accuracy in \% on the \tbd{738} real-task items; the second header row gives the number of items after
  the exclusions above. ``One
  pass'': items Qwen3.8-27B's direct readout answers correctly; ``Reasoning'': items it does not. J-: JevBench tiers.
  The reference rows define the split and the item selection, so their first two columns and the thinking row hold
  by construction. All models and all items are in Appendix~\ref{app:bench}.}
  \label{tab:real}
\end{table*}

\subsection{A reasoning backbone does not help}

We finally test whether a backbone trained to reason closes the gap. It does not. We train Kev with its released
recipe twice, once from Qwen3.5-Base, as released, and once from the post-trained Qwen3.5 that thinks
before answering, three seeds each (Table~\ref{tab:backbone} in Appendix~\ref{app:bench}). On the real-task items the two backbones are indistinguishable: Kev-4B reaches \tbd{87.2\%} and \tbd{87.4\%}, Kev-9B \tbd{90.1\%} and \tbd{90.1\%}, and on the items that need reasoning the difference is comparable to the spread across seeds. On the final-state items of Section~\ref{sec:diagnosis} up to 4k tokens, the reasoning backbone is worse: \tbd{70\%} vs.\
\tbd{50\%} at 4B and \tbd{76\%} vs.\ \tbd{71\%} at 9B for $k\geq2$. Whatever reasoning the backbone has learned
is exercised when it generates tokens; a model that answers in one forward pass does not draw on it.

On real tasks as on synthetic logs, then, Jev loses exactly the decisions that an LLM makes by writing
intermediate steps; a backbone trained to reason does not recover them. Section~\ref{sec:handoff} asks whether Jev
can tell when it meets such a decision.

%% file: figures/fig3_examples.tex
% Figure 3: two real-task items. Drawn in TikZ; included from sections/04_real.tex.
\begin{tikzpicture}[
    font=\scriptsize,
    x=1mm, y=1mm,
    lbl/.style={font=\scriptsize\sffamily\bfseries, anchor=north west, inner sep=0},
    meta/.style={font=\scriptsize\sffamily, text=black!55, anchor=north west, inner sep=0},
    state/.style={draw=black!35, anchor=north west, inner sep=1.2mm, text width=70.6mm, align=left, font=\tiny\ttfamily},
    txt/.style={anchor=north west, inner sep=0, align=left},
    opt/.style={draw=black!45, rounded corners=0.6mm, inner sep=0.8mm, font=\tiny\ttfamily, anchor=west,
      text height=1.6mm, text depth=0.45mm},
    ok/.style={opt, draw=tgtblue, line width=0.6pt, text=tgtblue},
    who/.style={anchor=west, inner sep=0, text width=26mm},
  ]
  \definecolor{tgtblue}{HTML}{2A78D6}
  \definecolor{errred}{HTML}{C0392B}

  % ---------- (a) an item that needs reasoning ----------
  \node[lbl] at (0,0) {(a) Needs reasoning};
  \node[meta] at (29,0) {JevBench-hard; demand \tbd{363} tokens};
  \node[state] (sa) at (0,-3.4) {A chemical batch may contain at most 0.080 percent impurity by mass; a reading exactly
    at the limit passes. A 2.50 kg sample contains 1.65 grams of impurity. Uncertainty is handled by adding 0.20 gram
    to the impurity mass before comparing; the total mass stays 2.50 kg.};
  \node[txt] (qa) at ([yshift=-1.6mm]sa.south west) {Q: classify the adjusted result.};
  \node[ok] (o1) at ([yshift=-3.6mm]qa.south west) {pass ($\leq$0.080\%)};
  \node[opt, right=0.8mm of o1] (o2) {narrow fail ($\leq$0.085\%)};
  \node[opt, right=0.8mm of o2] (o3) {clear fail};
  \node[who] (ja) at ([yshift=-3.6mm]o1.south west) {Jev 1.13.0};
  \node[anchor=west, inner sep=0] at (ja.east) {narrow fail, $p=0.46$\enspace\textcolor{errred}{wrong}};
  \node[who] (ta) at ([yshift=-3.4mm]ja.west) {Qwen3.8-27B, thinking};
  \node[anchor=west, inner sep=0] at (ta.east) {pass, 10/10 samples ($1.85/2500 = 0.074\%$)};

  % ---------- (b) an item with a disputed label ----------
  \node[lbl] (lb) at ([yshift=-4.5mm]ta.south west) {(b) Disputed label};
  \node[meta] at ([xshift=29mm]lb.north west) {Banking77; excluded};
  \node[state] (sb) at ([yshift=-3.4mm]lb.north west) {how many transactions can i make with a disposable card};
  \node[who] (gb) at ([yshift=-3.6mm]sb.south west) {gold label};
  \node[opt] at ([xshift=-0.8mm]gb.east) {get\_disposable\_virtual\_card};
  \node[who] (mb) at ([yshift=-4.2mm]gb.west) {Jev; thinking (10/10)};
  \node[opt] at ([xshift=-0.8mm]mb.east) {disposable\_card\_limits};
\end{tikzpicture}

%% file: sections/05_handoff.tex
\section{Does It Know, and Can It Hand Off?}
\label{sec:handoff}

\begin{figure*}[t]
  \centering
  \includegraphics[width=\textwidth]{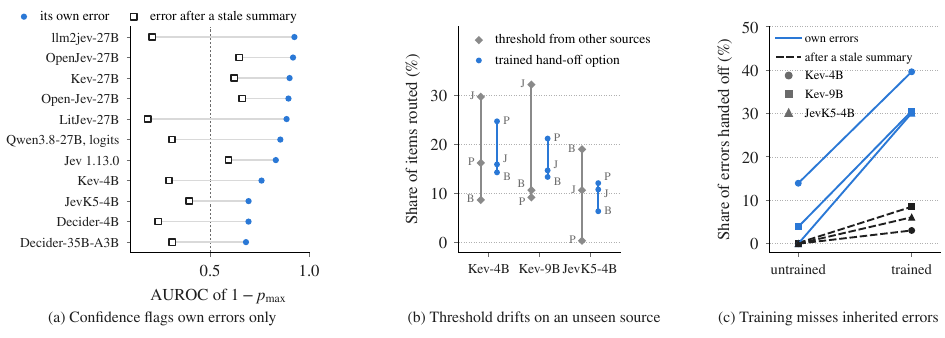}
  \caption{What confidence and a hand-off option catch. (a)~AUROC of $1-p_{\max}$ for a model's own final-state
  errors and for errors after a stale summary; synthetic logs up to 4k tokens; dashed line: chance. (b)~Share of
  real-task items routed on a held-out source (J: JevBench, P: PubMedQA, B: Banking77) by a threshold on the
  untrained model's confidence, chosen on the other two sources to match the trained option's rate there, and by
  the trained hand-off option. Banking77 is part of the Kev training data. (c)~Share of errors handed off on the
  synthetic logs before and after hand-off training. (b) and (c): mean over three seeds.}
  \label{fig:handoff}
\end{figure*}

A system built on Jev must send the items Jev would get wrong to a stronger model, either by reading Jev's
confidence $p_{\max}$ or by offering it an option to hand the question off.

\subsection{Confidence flags Jev's own errors, not inherited ones}

We first ask whether $p_{\max}$ separates Jev's wrong answers from its right ones. It does when Jev makes the error
itself: AUROC is \tbd{0.83} on the final-state items, with an expected calibration error of \tbd{0.03}, and
\tbd{0.94} on the real tasks. When the error comes from a stale summary, however, the signal is gone: AUROC is
\tbd{0.53}, and at $k=2$ it is \tbd{0.26}, so Jev is more confident when it follows the stale summary than when it
does not. Every model shows the same split (Figure~\ref{fig:handoff}a): on the synthetic logs its AUROC is
\tbd{0.68}--\tbd{0.93} for its own errors and \tbd{0.18}--\tbd{0.66} after a stale summary, below chance for
\tbd{7} of \tbd{11} models.

\subsection{Jev does not hand off}

We next offer an explicit way out, and Jev almost never takes it. We add one option, ``Cannot be determined
reliably from the state; hand this question to a stronger model'', and one sentence asking the model to choose it
when unsure. Jev chooses it for \tbd{1.6\%} of the real-task items and
\tbd{3} of its \tbd{40} errors, none of them among its \tbd{29} errors on JevBench-hard; on the synthetic logs it
hands off \tbd{3} of \tbd{145} wrong answers. The open models hand off more, but for every model except
Open-Jev-27B a confidence threshold that routes the same share of items leaves a more accurate remainder
(Table~\ref{tab:handoff}, top). The models carry the signal but do not act on it.

\subsection{Hand-off training helps where a threshold cannot be tuned}

We then ask whether the option can be trained in without new data, keeping the training data and number of steps
of the original recipes.
The training set is split in two halves; a model trained on one half answers the other, and every question it
answers wrongly is marked as unlearnable. In the final run each question carries the hand-off option with
probability 0.5, worded with paraphrases disjoint from the evaluation wording, and its target is the hand-off
option if the question is unlearnable and the answer otherwise. We train Kev-4B, Kev-9B and JevK5-4B, three seeds
each.

On the task the threshold was tuned for, training makes the option useful but not better than the threshold
(Table~\ref{tab:handoff}, bottom). Accuracy without the option is unchanged, and the share of errors handed off
doubles, from \tbd{23\%} to \tbd{47\%} for Kev-4B and from \tbd{22\%} to \tbd{44\%} for
Kev-9B. A threshold on the untrained model that routes the same share of items is still more accurate, by
\tbd{1.3}--\tbd{1.7} points.

Without labelled data from the target task, however, the threshold cannot be set, and the trained option is the
steadier signal (Figure~\ref{fig:handoff}b). Choosing the threshold on two sources so that it routes the same share as
the trained model and applying it to the third routes \tbd{9}--\tbd{32\%} of the items for Kev-4B and Kev-9B instead of about
\tbd{15\%}; for JevK5-4B on PubMedQA, whose yes/no questions never give $p_{\max}<0.5$, it routes none and catches \tbd{2\%} of the errors,
where the trained option catches \tbd{71\%}. The trained option routes \tbd{6}--\tbd{25\%} on every source
and needs no tuning.

Neither mechanism reaches the inherited errors (Figure~\ref{fig:handoff}c). On the synthetic logs up to 4k
tokens, training raises the share of final-state errors that the models hand off from \tbd{0}--\tbd{14\%} to
\tbd{30}--\tbd{40\%}, but the share of errors caused by a stale summary only from \tbd{0\%} to
\tbd{3}--\tbd{9\%}: the training labels mark questions the model cannot answer, not inputs that are wrong. In
our experiments only generated reasoning checks the input: with thinking, Qwen3.8-27B rejects \tbd{95\%} of the
stale summaries (Table~\ref{tab:crossmodel}).

\begin{table}[t]
  \centering
  \small
  \setlength{\tabcolsep}{3.5pt}
  \begin{tabular}{@{}l c cc cc@{}}
    \toprule
    & & \multicolumn{2}{c}{\textbf{Hand-off}} & \multicolumn{2}{c}{\textbf{Acc.\ of the rest}} \\
    \cmidrule(lr){3-4}\cmidrule(l){5-6}
    \textbf{Model} & Acc. & Rate & Errors & Option & Threshold \\
    \midrule
    Jev 1.13.0          & 94.6 & 1.6 & 8 & 94.5 & 95.5 \\
    Kev-27B             & 93.9 & 9.6 & 20 & 93.9 & 97.0 \\
    LitJev-27B          & 91.3 & 3.1 & 16 & 92.3 & 93.6 \\
    llm2jev-27B         & 92.0 & 3.8 & 14 & 93.5 & 94.1 \\
    JevK5-9B            & 90.2 & 3.1 & 10 & 90.8 & 92.6 \\
    Open-Jev-27B        & 80.8 & 9.6 & 25 & 88.8 & 82.6 \\
    \midrule
    Kev-4B (ours)       & 87.1 & 10.1 & 23 & 89.0 & 91.5 \\
    \quad + hand-off & 87.2 & 17.0 & 47 & 92.1 & 93.6 \\
    Kev-9B (ours)       & 90.1 & 9.3 & 22 & 91.5 & 93.8 \\
    \quad + hand-off & 90.0 & 14.6 & 44 & 93.8 & 95.1 \\
    JevK5-4B (ours)     & 88.5 & 7.2 & 22 & 89.9 & 92.3 \\
    \quad + hand-off & 88.6 & 9.9 & 35 & 91.5 & 93.2 \\
    \bottomrule
  \end{tabular}
  \caption{Hand-off on the \tbd{738} real-task items; all columns in \%. Acc.: accuracy without the hand-off option. Rate: share of
  items handed off when the option is offered; Errors: share of the model's errors (without the option) that it
  hands off. Acc.\ of the rest: accuracy on the items not handed off, by the option or by a threshold on the
  confidence of the untrained model that routes the same share. Bottom: our training runs, mean over three seeds; ``(ours)'' reproduces the released recipe and ``+ hand-off'' adds
  hand-off training. Kev-0.8B is in Appendix~\ref{app:bench}.}
  \label{tab:handoff}
\end{table}

Jev therefore does not recognise its limit reliably, and least of all when the error comes from its input. The
division of labour for a system built on Jev follows: where labelled data from the task exist, a threshold on
$p_{\max}$ routes Jev's own errors best; where they do not, the trained hand-off option is more robust; and
decisions that need intermediate steps, above all checking an input Jev did not compute itself, have to be made by
a model that writes them.

%% file: sections/06_related.tex
\section{Related Work}
\label{sec:related}

\paragraph{Decision models.}
Jev~\citep{typesafe2026jev} introduced the interface we study; its replicas fine-tune an LLM, wrap one, or read
its option logits (Section~\ref{sec:models}). Reading option scores from an LLM is also how multiple-choice
benchmarks are commonly evaluated, and such scores depend on the prompt, the option labels and their
order~\citep{zhao2021calibrate,zheng2024large}. Public evaluations of decision models report accuracy and
calibration over mixed tasks~\citep{jevbench,jevals}; we instead isolate the inputs on which they fail.

\paragraph{Long contexts and state tracking.}
Retrieval from long contexts degrades with length, the position of the target and the difficulty of the
task~\citep{kamradt2023needle,liu2024lost,hsieh2024ruler}, and
reasoning over facts spread through them degrades further~\citep{kuratov2024babilong}. Language models
track entity states only partly~\citep{kim2023entity}, and recalling the latest value of a key suffers from
interference by its earlier values~\citep{wang2025unable}. Decision models retrieve without error up to their
limit, and in our logs the target's own earlier updates are harmless (R1); the failure comes from similar records
of other users.

\paragraph{What one forward pass can compute.}
Writing intermediate steps improves LLM reasoning~\citep{wei2022chain}, and theory shows why: bounded-depth
transformers provably cannot solve some problems without them~\citep{feng2023towards,merrill2024expressive},
state tracking among them~\citep{merrill2024illusion}; LLMs also complete multi-hop inferences only partly within
one pass~\citep{yang2024latent}. Decision models give these steps up by design; we measure what this costs on
practical decisions, on synthetic logs as well as on real tasks.

\paragraph{Knowing when to abstain.}
Large LLMs are well calibrated on multiple-choice questions in a suitable format~\citep{kadavath2022language},
and a threshold on the maximum probability is the standard way to abstain~\citep{geifman2017selective}. Learning to defer trains a model to pass cases to an
expert~\citep{mozannar2020consistent}, and refusal tuning labels questions beyond the model's knowledge as ones to
decline~\citep{zhang2024rtuning}, much as our hand-off training labels questions a held-out model answers wrongly.
In our experiments both catch only the errors a model makes itself: like LLMs that adopt coherent but conflicting
context~\citep{xie2024adaptive}, Jev adopts a stale summary, although it contradicts the log rather than the
model's memory, and neither mechanism detects it.

%% file: sections/07_conclusion.tex
\section{Conclusion}
\label{sec:conclusion}

We asked what Jev cannot decide, and found a single limit: Jev, like the open Jev-style models we evaluated, cannot
make a decision that an LLM makes only by writing out intermediate steps. These models find and select records
reliably, but they fail to track a state among similar records, follow a precomputed state without checking it, and
lose the real decision items that need reasoning, the more so the more reasoning an item needs. Their confidence
flags the errors they make themselves but not those they inherit from a wrong input, and neither an explicit nor a
trained hand-off option closes this gap. One-pass decision models are therefore best used as fast selectors over a
state that a reasoning model has computed and checked. Whether training on such decisions, or a few internal steps
before the readout, can move the limit remains open.

%% file: sections/appendix.tex
\clearpage
\section{Models and Serving Details}
\label{app:models}

\subsection{Decision models}

Table~\ref{tab:serving} lists every model we evaluate. Each model is served by the code its authors released, behind
an endpoint that accepts a state, a question and an option list and returns a probability per option; the two
exceptions are the letter-logit readout of plain Qwen models, which we implement ourselves, and CLM, which scores
options by embedding similarity. All models receive the same items with the same option order. An input that
exceeds a model's limit is recorded as unsupported and never truncated: Decider is started with a state limit above
its 32,768-token row limit so that it refuses instead of cutting, and CLM is run with its truncation disabled
(\texttt{--max-tokens 0}). Requests that fail with HTTP 429, 503 or 529 are retried with exponential backoff; any
other failure is recorded as an error and excluded from accuracy.

For the open checkpoints we pin the Hugging Face revision. The revisions behind our numbers are
\texttt{139fdd94} (Kev-4B), \texttt{2629c06a} (Kev-9B), \texttt{01b81998} (Kev-27B, the snapshot with a separate
adapter) and \texttt{27d2d6b8} (JevK5-4B v0.2). The \texttt{main} branch of the Kev repositories was replaced on
2026-10-10; the new Kev-9B weights differ from the ones we report (the same answer on 82\% of the final-state items
up to 4k tokens),
whereas the new Kev-4B weights are identical to ours.

\paragraph{Letter-logit readout.}
For plain Qwen models (Qwen3.5-4B and 9B, each with its Base model, and Qwen3.8-27B) we follow the recipe of
jqv~\citep{jqv} and SemIf~\citep{semif}. The prompt, in the Qwen chat format with thinking disabled, contains the
document, the question and the options labelled A, B, \dots, asks for the letter only, and ends with
\texttt{Answer:}. The next-token probabilities of the offered letters, renormalised, are the answer distribution;
nothing is generated and no temperature is fitted. Items with more than 26 options are unsupported, which removes the
32--64-option selection cells.

\paragraph{Thinking reference.}
Qwen3.8-27B with thinking is served with SGLang and queried through the OpenAI-compatible chat API with the sampling
settings of its model card (temperature 1.0, top-$p$ 0.95, top-$k$ 20) and at most 24,000 new tokens; a reply cut
off at that limit counts as wrong. The answer is parsed from the reply, so it has no probabilities and no AUROC.

\begin{table*}[t]
  \centering
  \small
  \setlength{\tabcolsep}{4pt}
  \begin{tabularx}{\textwidth}{@{}>{\raggedright\arraybackslash}p{0.27\textwidth} >{\raggedright\arraybackslash}p{0.2\textwidth}
    >{\raggedright\arraybackslash}X >{\raggedright\arraybackslash}p{0.17\textwidth}@{}}
    \toprule
    \textbf{Model} & \textbf{Backbone} & \textbf{How the answer is read} & \textbf{Input limit} \\
    \midrule
    Jev 1.13.0~\citep{typesafe2026jev} & undisclosed & undisclosed (hosted API) & 32k, state + longest question \\
    \midrule
    Kev 0.8B, 4B, 9B, 27B~\citep{kev} & Qwen3.5-Base; Qwen3.8-27B & LoRA and a pointer head over the options & 4B, 27B: $<$8k \\
    JevK5 4B v0.2, v0.3; 9B v0.3.3~\citep{jevk5} & Qwen3.5-4B, 9B & merged LoRA, option-letter logits & -- \\
    Decider 0.8B--35B-A3B~\citep{decider} & Qwen3.5-Base & fine-tuned, one forward pass & 32,768 per row \\
    Open-Jev-27B v1.1~\citep{openjev27b} & Qwen3.8-27B & fine-tuned, one forward pass & 16,384 \\
    OpenJev-27B~\citep{openjev} & Qwen3.8-27B & fine-tuned, one forward pass & 16,384; $\leq$52 options \\
    \midrule
    LitJev-27B~\citep{litjev} & Qwen3.8-27B & training-free decision layer & 16,384 \\
    llm2jev 4B, 27B~\citep{llm2jev} & Qwen3.5-4B; Qwen3.8-27B & label-token log-probabilities after one prefill & -- \\
    OpenJev-SGLang~\citep{openjevsglang} & Qwen3.6-35B-A3B (bf16) & label-token log-probabilities after one prefill & 32,768 \\
    Qwen, letter logits (ours) & Qwen3.5-4B, 9B; Qwen3.8-27B & option-letter logits after one forward pass & 32,768; $\leq$26 options \\
    \midrule
    CLM-v0.1-8B~\citep{clm} & Qwen3-8B (frozen) & similarity of contrastive state and option embeddings & 40,960 \\
    \bottomrule
  \end{tabularx}
  \caption{Models and how they are served; repositories are given in the references. Input limit: the largest
  input the model accepts in our runs, in tokens of its own tokenizer; ``--'': no input in our benchmarks exceeded
  it. OpenJev-SGLang is run with the unquantised Qwen3.6-35B-A3B~\citep{qwen36} rather than the NVFP4 default of its
  repository.}
  \label{tab:serving}
\end{table*}

\subsection{Real-task items and reasoning demand}

JevBench items are the public easy, original and hard splits of the repository~\citep{jevbench}. Jevals items are
rebuilt from the pinned Hugging Face rows of suite 0.1.0~\citep{jevals} exactly as the Jevals harness builds them,
and each state is checked against the SHA-256 hash in the suite. Every item is reduced to a closed option list:
yes/no questions get the options yes and no, with the criteria as their descriptions, and score questions their
levels.

Reasoning demand is measured with Qwen3.8-27B served by SGLang. The direct readout disables thinking, forces the
reply to begin with \texttt{Answer:}, and scores each option label by the log-probability of its tokens followed by
the end of the label, normalised over the options; nothing is generated. The no-thinking and thinking modes draw
10 samples each with the sampling settings of the model card (no thinking: temperature 0.7, top-$p$ 0.8, top-$k$ 20,
presence penalty 1.5; thinking: temperature 1.0, top-$p$ 0.95, top-$k$ 20) and at most 4,096 and 32,768 new tokens;
a truncated sample counts as wrong. The demand of an item is the median number of thinking tokens over the thinking
samples that answer it correctly.

\subsection{Training runs}

\paragraph{Kev.}
All Kev runs use the decision-v7 configuration released with Kev: the decision-v7 training suite with at most 1,000
records per source, two epochs, learning rate $5\times10^{-5}$ ($10^{-4}$ at 0.8B), LoRA rank 16, an effective batch
of 8, and a pinned base revision; seeds 2, 3 and 4. The backbone comparison of Section~\ref{sec:real} changes only the
base model, from Qwen3.5-4B-Base and Qwen3.5-9B-Base to the post-trained Qwen3.5-4B and Qwen3.5-9B~\citep{qwen35}.

\paragraph{JevK5.}
The JevK5 runs reproduce the v0.2 recipe. Its teacher-written questions are not released, so we regenerate them with
the released teacher script and settings and add the same public training splits; we then train with LoRA rank 16,
learning rate $3\times10^{-5}$,
two epochs, inputs up to 2,048 tokens, 20 warm-up steps, and a softmax temperature fitted after training; seeds 0, 1
and 2.

\paragraph{Hand-off training.}
The training records are split into two halves (6,334 and 6,242 records for Kev; 3,272 each for JevK5). A model
trained with the unchanged recipe on one half answers every question of the other half, and a question it answers
wrongly is marked unlearnable. A separate pair of models is trained for each backbone, so the labels are the
backbone's own: 2,220 Kev records contain an unlearnable question for the 4B labels, 2,069 for 9B and 2,939 for
0.8B, and 1,331 of the 6,544 JevK5 training items are marked unlearnable. The final run sees each question once per epoch; with
probability 0.5 it is rewritten as a choice with an added hand-off option at a random position and a sentence asking
the model to choose it when unsure. Option text and sentence are drawn from six and four paraphrases that exclude the
evaluation wording. The target is the hand-off option for an unlearnable question and the gold answer otherwise.
Everything else, including the number of steps, follows the recipe above.

\section{Full Benchmark Results}
\label{app:bench}

\subsection{Controlled benchmark}

Table~\ref{tab:fullbench} extends Table~\ref{tab:crossmodel} to every model. Jev answers every retrieval item
correctly in both haystacks, at every depth and up to 28k tokens of essay text, and every selection item with 2 to
64 options; its final-state accuracy by context length is shown in Figure~\ref{fig:heatmap}. Two models fail before
the final-state question. Kev-0.8B already loses event lookup (\tbd{66\%}). CLM-v0.1-8B, which scores options by
embedding similarity, is at chance on retrieval with four options (\tbd{25\%}) and answers no single-update
final-state item; it accepts every summary, correct or stale, and its \tbd{80\%} at $k=16$ next to \tbd{0\%} at
$k=1$ points to a preference for particular options rather than to reading the log. Several models reach a high rejection rate for stale summaries only by
rejecting nearly every summary: Qwen3.5-4B-Base rejects \tbd{98\%} of stale summaries and accepts \tbd{2\%} of correct
ones.

\begin{table*}[t]
  \centering
  \small
  \setlength{\tabcolsep}{3pt}
  \resizebox{\textwidth}{!}{%
  \begin{tabular}{@{}l cccc ccc cc cc cc@{}}
    \toprule
    & \multicolumn{4}{c}{\textbf{Find and select}} & \multicolumn{3}{c}{\textbf{Final state}} & \multicolumn{2}{c}{\textbf{With summary}}
    & \multicolumn{2}{c}{\textbf{Verify}} & \multicolumn{2}{c}{\textbf{AUROC}} \\
    \cmidrule(lr){2-5}\cmidrule(lr){6-8}\cmidrule(lr){9-10}\cmidrule(lr){11-12}\cmidrule(l){13-14}
    \textbf{Model} & Retr. & Sel.\,$\leq$26 & Sel.\,32--64 & Event & $k{=}1$ & $k{\geq}2$ & $k{=}16$ & correct & stale
    & accept & reject & own & stale \\
    \midrule
    Jev 1.13.0                         & 100 & 100 & 100 & 100 & 100 & 84 & 64 & 100 & 58 & 92 & 70 & 0.83 & 0.57 \\
    \midrule
    Kev-27B$^\dagger$                  & 100 & 100 & 100 & 100 & 100 & 91 & 88 & 100 & 60 & 100 & 62 & 0.90 & 0.62 \\
    Kev-9B                             & 100 & 100 & 100 & 100 & 100 & 70 & 54 & 100 & 55 & 81 & 26 & 0.70 & 0.38 \\
    Kev-4B$^\dagger$                   & 100 & 100 & 100 & 100 & 98 & 66 & 41 & 100 & 53 & 88 & 38 & 0.76 & 0.29 \\
    Kev-0.8B                           & 86 & 99 & 97 & 66 & 18 & 32 & 18 & 91 & 25 & 66 & 41 & 0.56 & 0.46 \\
    JevK5-4B v0.2                      & 100 & 100 & 100 & 99 & 98 & 53 & 56 & 100 & 47 & 73 & 52 & 0.69 & 0.38 \\
    JevK5-4B v0.3                      & 100 & 100 & 99 & 99 & 100 & 61 & 65 & 99 & 55 & 84 & 32 & 0.75 & 0.49 \\
    JevK5-9B v0.3.3                    & 100 & 100 & 100 & 100 & 100 & 68 & 86 & 98 & 64 & 92 & 26 & 0.79 & 0.50 \\
    Decider-35B-A3B                    & 100 & 100 & 100 & 99 & 83 & 65 & 60 & 100 & 39 & 85 & 31 & 0.70 & 0.27 \\
    Decider-4B                         & 100 & 100 & 100 & 98 & 100 & 67 & 80 & 100 & 39 & 100 & 0 & 0.71 & 0.21 \\
    Decider-2B                         & 100 & 100 & 100 & 98 & 95 & 69 & 95 & 100 & 42 & 85 & 21 & 0.67 & 0.63 \\
    Decider-0.8B                       & 97 & 100 & 100 & 82 & 67 & 54 & 68 & 100 & 37 & 89 & 17 & 0.63 & 0.59 \\
    Open-Jev-27B v1.1                  & 100 & 100 & 100 & 100 & 100 & 89 & 78 & 100 & 72 & 96 & 74 & 0.87 & 0.58 \\
    OpenJev-27B                        & 100 & 100 & 100 & 100 & 100 & 95 & 93 & 100 & 66 & 98 & 56 & 0.87 & 0.62 \\
    \midrule
    LitJev-27B                         & 100 & 100 & 100 & 100 & 98 & 79 & 76 & 100 & 41 & 100 & 7 & 0.86 & 0.16 \\
    llm2jev-27B                        & 100 & 100 & 100 & 100 & 100 & 86 & 81 & 100 & 43 & 51 & 73 & 0.90 & 0.20 \\
    llm2jev-4B                         & 100 & 100 & 77 & 99 & 83 & 37 & 25 & 100 & 38 & 10 & 89 & 0.70 & 0.23 \\
    OpenJev-SGLang-35B-A3B             & 100 & 100 & 98 & 100 & 87 & 50 & 61 & 100 & 38 & 59 & 54 & 0.74 & 0.24 \\
    \midrule
    Qwen3.8-27B                        & 100 & 100 & -- & 100 & 100 & 77 & 77 & 100 & 42 & 51 & 74 & 0.82 & 0.28 \\
    Qwen3.5-9B                         & 100 & 100 & -- & 100 & 98 & 58 & 79 & 99 & 40 & 98 & 5 & 0.74 & 0.26 \\
    Qwen3.5-9B-Base                    & 100 & 100 & -- & 100 & 95 & 69 & 83 & 100 & 48 & 53 & 52 & 0.77 & 0.39 \\
    Qwen3.5-4B                         & 100 & 100 & -- & 100 & 88 & 39 & 30 & 100 & 41 & 10 & 88 & 0.68 & 0.22 \\
    Qwen3.5-4B-Base                    & 100 & 100 & -- & 99 & 83 & 63 & 79 & 100 & 49 & 2 & 98 & 0.76 & 0.39 \\
    \midrule
    CLM-v0.1-8B                        & 25 & 36 & 10 & 26 & 0 & 50 & 80 & 24 & 24 & 100 & 0 & 0.62 & 0.42 \\
    \midrule
    Qwen3.8-27B, thinking$^\ddagger$   & -- & -- & -- & 100 & 100 & 96 & 100 & 100 & 83 & 99 & 95 & -- & -- \\
    \bottomrule
  \end{tabular}}
  \caption{Controlled benchmark, all models, contexts up to 8k tokens; accuracies and shares in \%, AUROC in
  $[0,1]$. Groups: Jev; fine-tuned; training-free; plain
  Qwen with the letter-logit readout (no items with more than 26 options); contrastive. Verify: share of correct
  summaries accepted and of stale summaries rejected. AUROC of $1-p_{\max}$ for the model's own final-state errors and
  for errors after a stale summary. $^\dagger$Up to 4k tokens. $^\ddagger$Generates its reasoning; state-tracking
  items only, no probabilities.}
  \label{tab:fullbench}
\end{table*}

\subsection{Prompt and log variants}

The variants of Table~\ref{tab:prompt} are built from each final-state item without changing its options or their
order. P1--P3 append one sentence to the question. R1 keeps only the log lines that mention the target user. R2
reverses the log and replaces ``After the last log entry'' in the question by ``The log is listed from the newest
entry to the oldest. After the most recent log entry''. R3 moves the target user's lines, in order, to the end of the
log and keeps every other line. R4 replaces the other users' lines of R3 by paragraphs of the essay corpus with the
same total length in tokens. R3 and R4 therefore have the length of the original item, and the last $k$ lines of
both are exactly the target user's updates. Table~\ref{tab:prompt9b} gives Kev-9B, which follows Kev-4B.

\begin{table}[t]
  \centering
  \small
  \begin{tabular}{@{}l cc@{}}
    \toprule
    \textbf{Variant} & $k\geq2$ & $k=16$ \\
    \midrule
    P0 original          & 69 & 48 \\
    \midrule
    P1 step by step      & 69 & 56 \\
    P2 procedure         & 69 & 56 \\
    P3 name failure      & 67 & 48 \\
    \midrule
    R1 target only       & 71 & 50 \\
    R2 newest first      & 20 & 7 \\
    R3 target at end     & 66 & 44 \\
    R4 R3 + essays       & 90 & 83 \\
    \bottomrule
  \end{tabular}
  \caption{Final-state accuracy in \% of Kev-9B under the variants of Table~\ref{tab:prompt}, up to 4k tokens.}
  \label{tab:prompt9b}
\end{table}

\subsection{Real tasks}

Table~\ref{tab:realfull} gives every model on every source before any exclusion, the 831 items without HelpSteer2,
and the 738 items of Section~\ref{sec:real}. On HelpSteer2 every item asks for one of five helpfulness levels, and
always answering the most frequent level (4) is correct for \tbd{42\%} of the items; no model, with or without
thinking, is clearly above that. The 93 excluded items are those Qwen3.8-27B with thinking answers correctly in at
most 5 of 10 samples: \tbd{65} from Banking77, \tbd{25} from PubMedQA and \tbd{3} from JevBench-hard.

\begin{table*}[t]
  \centering
  \small
  \begin{tabular}{@{}l cccccc cc@{}}
    \toprule
    & \multicolumn{3}{c}{\textbf{JevBench}} & & & & \multicolumn{2}{c}{\textbf{Pooled}} \\
    \cmidrule(lr){2-4}\cmidrule(l){8-9}
    \textbf{Model} & easy & original & hard & PubMedQA & Banking77 & HelpSteer2 & 831 items & 738 items \\
    & 48 & 72 & 111 & 300 & 300 & 300 & & \\
    \midrule
    Jev 1.13.0           & 100.0 & 98.6 & 71.2 & 92.0 & 80.3 & 41.7 & 86.0 & 94.6 \\
    Kev-27B              & 100.0 & 100.0 & 70.3 & 89.7 & 83.3 & 44.3 & 86.3 & 93.9 \\
    Kev-4B               & 100.0 & 93.1 & 51.4 & 83.7 & 82.7 & 44.0 & 80.7 & 86.4 \\
    JevK5-4B v0.2        & 100.0 & 94.4 & 73.9 & 85.0 & 69.7 & 35.7 & 79.7 & 87.7 \\
    JevK5-4B v0.3        & 100.0 & 94.4 & 74.8 & 83.7 & 71.7 & 46.3 & 80.0 & 88.1 \\
    JevK5-9B v0.3.3      & 100.0 & 95.8 & 76.6 & 88.3 & 72.7 & 47.0 & 82.4 & 90.2 \\
    Decider-4B           & 100.0 & 98.6 & 64.0 & 87.0 & 85.0 & 49.0 & 85.0 & 90.4 \\
    Open-Jev-27B v1.1    & 100.0 & 95.8 & 73.0 & 71.7 & 65.3 & 39.3 & 73.3 & 80.8 \\
    LitJev-27B           & 100.0 & 98.6 & 73.9 & 88.0 & 74.3 & 36.0 & 82.8 & 91.3 \\
    llm2jev-27B          & 100.0 & 95.8 & 72.1 & 88.7 & 75.7 & 30.7 & 83.0 & 92.0 \\
    llm2jev-4B           & 100.0 & 76.4 & 59.5 & 84.7 & 67.7 & 32.3 & 75.3 & 82.0 \\
    \midrule
    Qwen3.8-27B, direct  & 97.9 & 94.4 & 66.7 & 90.3 & 67.7 & 36.3 & 79.8 & 87.5 \\
    Qwen3.8-27B, no thinking & 100.0 & 98.6 & 97.6 & 89.0 & 78.0 & 34.0 & 87.6 & 96.9 \\
    Qwen3.8-27B, thinking & 100.0 & 99.7 & 97.3 & 91.6 & 78.6 & 32.7 & 88.9 & 99.0 \\
    \bottomrule
  \end{tabular}
  \caption{Real-task accuracy in \% before exclusion; the second header row gives the number of items. 831 items: all
  sources except HelpSteer2. 738 items: additionally without the 93 disputed items (Section~\ref{sec:real}). The
  sampled modes are averaged over 10 samples.}
  \label{tab:realfull}
\end{table*}

\subsection{Reasoning backbone}

Table~\ref{tab:backbone} gives the backbone comparison of Section~\ref{sec:real}.

\begin{table}[t]
  \centering
  \small
  \setlength{\tabcolsep}{2pt}
  \begin{tabular}{@{}ll ccc c@{}}
    \toprule
    & & \multicolumn{3}{c}{\textbf{Real tasks}} & \textbf{Final state} \\
    \cmidrule(lr){3-5}\cmidrule(l){6-6}
    & \textbf{Backbone} & All & One pass & Reasoning & $k\geq2$ \\
    \midrule
    \multirow{2}{*}{4B} & Base      & 87.2\,{\scriptsize$\pm$0.5} & 91.8\,{\scriptsize$\pm$0.4} & 54\,{\scriptsize$\pm$3} & 70\,{\scriptsize$\pm$5} \\
                        & Reasoning & 87.4\,{\scriptsize$\pm$0.4} & 91.4\,{\scriptsize$\pm$0.3} & 58\,{\scriptsize$\pm$1} & 50\,{\scriptsize$\pm$3} \\
    \midrule
    \multirow{2}{*}{9B} & Base      & 90.1\,{\scriptsize$\pm$0.2} & 94.1\,{\scriptsize$\pm$0.2} & 60\,{\scriptsize$\pm$3} & 76\,{\scriptsize$\pm$0} \\
                        & Reasoning & 90.1\,{\scriptsize$\pm$0.5} & 94.2\,{\scriptsize$\pm$0.5} & 59\,{\scriptsize$\pm$2} & 71\,{\scriptsize$\pm$2} \\
    \bottomrule
  \end{tabular}
  \caption{Kev trained with the same recipe from Qwen3.5-Base or from the post-trained, thinking Qwen3.5; accuracy
  in \%, mean $\pm$ standard deviation over three seeds. Real tasks: the \tbd{738} items, restricted to those every seed of a
  size answers (\tbd{717} at 4B, \tbd{712} at 9B). Final state: Section~\ref{sec:diagnosis}, up to 4k tokens.}
  \label{tab:backbone}
\end{table}

\subsection{Hand-off}

Kev-0.8B, trained and evaluated like the larger Kev models in Table~\ref{tab:handoff}, follows the same pattern
with larger effects. Hand-off training lowers accuracy without the option from \tbd{79.3\%} to \tbd{77.9\%}, raises the
share of items handed off from \tbd{7.7\%} to \tbd{32.8\%} and the share of errors handed off from \tbd{15\%} to
\tbd{58\%}. The items it answers are then correct \tbd{86.5\%} of the time, against \tbd{91.8\%} for a threshold on the
untrained model that routes the same share.

%% file: custom.bib
@misc{typesafe2026jev,
    author = {{TypeSafe AI}},
    title  = {Jev: typed decisions in one forward pass ({API} model jev-1.13.0)},
    year   = {2026},
    url    = {https://docs.typesafe.ai/models}
}

@misc{kev,
    author       = {Palmer, Jared},
    title        = {{Kev-4B}: a calibrated decision model on {Qwen3.5-4B}},
    year         = {2026},
    howpublished = {\url{https://huggingface.co/jaredpalmer/kev-4b}},
    note         = {Kev 1.0; code at \url{https://github.com/jaredpalmer/kev}}
}

@misc{jevk5,
    author       = {{alibiserikbay}},
    title        = {{JevK5}: an open-source alternative to {TypeSafe's} {Jev} for typed decisions},
    year         = {2026},
    howpublished = {\url{https://huggingface.co/alibiserikbay/JevK5}},
    note         = {Runtime at \url{https://github.com/allebee/jevk5}}
}

@misc{decider,
    author       = {{Mapika}},
    title        = {decider: typed decisions with calibrated probabilities in one forward pass},
    year         = {2026},
    howpublished = {\url{https://huggingface.co/Mapika/decider-4b}},
    note         = {Code at \url{https://github.com/Mapika/decider}}
}

@misc{openjev27b,
    author       = {{Zefan-Cai}},
    title        = {{Open-Jev}},
    year         = {2026},
    howpublished = {\url{https://github.com/Zefan-Cai/Open-Jev}},
    note         = {Checkpoint \url{https://huggingface.co/ZefanCai/Open-Jev-27B-v1.1}}
}

@misc{openjev,
    author       = {{OpenJev}},
    title        = {{OpenJev}},
    year         = {2026},
    howpublished = {\url{https://huggingface.co/openjev/openjev}}
}

@misc{litjev,
    author       = {{ZhengxuYu}},
    title        = {{LitJev}: A {JEV}-like Decision Layer for Off-the-Shelf {LLMs}},
    year         = {2026},
    howpublished = {\url{https://github.com/zhengxuyu/litjev}},
    note         = {Version 0.1.0}
}

@misc{llm2jev,
    author       = {{tic-top}},
    title        = {llm2jev: any chat model, any engine as a {Jev}-compatible probability decision service},
    year         = {2026},
    howpublished = {\url{https://github.com/tic-top/llm2jev}}
}

@misc{semif,
    author       = {{TheoLeeCJ}},
    title        = {{SemIf}: semantic if-decisions from open models},
    year         = {2026},
    howpublished = {\url{https://github.com/TheoLeeCJ/SemIf}},
    note         = {Formerly OpenJev}
}

@misc{clm,
    author = {Jacky Kwok and Hangoo Kang and Tarun Suresh and Jon Saad-Falcon and Marco Pavone and Christopher R{\'e}
              and Azalia Mirhoseini},
    title  = {Contrastive Language Models: A System One Model for Fast and Generalizable Decision-Making},
    year   = {2026},
    note   = {Notion Blog},
    url    = {https://contrastive-lm.notion.site}
}

@article{qwen3,
    author  = {{Qwen Team}},
    title   = {{Qwen3} Technical Report},
    journal = {arXiv preprint arXiv:2505.09388},
    year    = {2025}
}

@misc{jevbench,
    author       = {{Benchmark Heaven}},
    title        = {{JevBench}: Benchmark for {AI} Decision Models},
    year         = {2026},
    howpublished = {\url{https://github.com/fstandhartinger/jevbench}}
}

@misc{jevals,
    author       = {{Jevals}},
    title        = {{Jevals}: independent benchmark data for {Jev} and {Jev}-type models},
    year         = {2026},
    howpublished = {\url{https://github.com/Jevals/jevals-data}},
    note         = {Release 2026-09-18, suite 0.1.0, CC-BY-4.0; \url{https://jevals.com}}
}

@inproceedings{jin2019pubmedqa,
    author    = {Jin, Qiao and Dhingra, Bhuwan and Liu, Zhengping and Cohen, William and Lu, Xinghua},
    title     = {{PubMedQA}: A Dataset for Biomedical Research Question Answering},
    booktitle = {Proceedings of the 2019 Conference on Empirical Methods in Natural Language Processing and the 9th
                 International Joint Conference on Natural Language Processing (EMNLP-IJCNLP)},
    pages     = {2567--2577},
    year      = {2019}
}

@inproceedings{casanueva2020banking77,
    author    = {Casanueva, I{\~n}igo and Tem{\v{c}}inas, Tadas and Gerz, Daniela and Henderson, Matthew and Vuli{\'c}, Ivan},
    title     = {Efficient Intent Detection with Dual Sentence Encoders},
    booktitle = {Proceedings of the 2nd Workshop on Natural Language Processing for Conversational AI},
    pages     = {38--45},
    year      = {2020}
}

@article{wang2024helpsteer2,
    author  = {Wang, Zhilin and Dong, Yi and Delalleau, Olivier and Zeng, Jiaqi and Shen, Gerald and Egert, Daniel and
               Zhang, Jimmy J. and Sreedhar, Makesh Narsimhan and Kuchaiev, Oleksii},
    title   = {{HelpSteer2}: Open-source dataset for training top-performing reward models},
    journal = {arXiv preprint arXiv:2406.08673},
    year    = {2024}
}

@misc{jqv,
    author       = {{Octalab-Inc}},
    title        = {jqv: Decision {API} on stock {Qwen3}},
    year         = {2026},
    howpublished = {\url{https://github.com/Octalab-Inc/jqv}}
}

@misc{openjevsglang,
    author       = {{ekzhang}},
    title        = {openjev-sglang: {Jev}-compatible {API} endpoint based on open models (prefill-only)},
    year         = {2026},
    howpublished = {\url{https://github.com/ekzhang/openjev-sglang}}
}

@misc{qwen35,
    author = {{Qwen Team}},
    title  = {{Qwen3.5}: Towards Native Multimodal Agents},
    month  = {February},
    year   = {2026},
    url    = {https://qwen.ai/blog?id=qwen3.5}
}

@misc{qwen36,
    author = {{Qwen Team}},
    title  = {{Qwen3.6-35B-A3B}: Agentic Coding Power, Now Open to All},
    month  = {April},
    year   = {2026},
    url    = {https://qwen.ai/blog?id=qwen3.6-35b-a3b}
}

@misc{qwen38,
    author = {{Qwen Team}},
    title  = {{Qwen3.8-Max}: A New Bar for Coding and Cowork},
    month  = {August},
    year   = {2026},
    url    = {https://qwen.ai/blog?id=qwen3.8}
}

@misc{kamradt2023needle,
    author       = {Gregory Kamradt},
    title        = {Needle In A Haystack -- pressure testing {LLMs}},
    year         = {2023},
    howpublished = {\url{https://github.com/gkamradt/LLMTest_NeedleInAHaystack}}
}

@article{liu2024lost,
    author  = {Liu, Nelson F. and Lin, Kevin and Hewitt, John and Paranjape, Ashwin and Bevilacqua, Michele and
               Petroni, Fabio and Liang, Percy},
    title   = {Lost in the Middle: How Language Models Use Long Contexts},
    journal = {Transactions of the Association for Computational Linguistics},
    volume  = {12},
    pages   = {157--173},
    year    = {2024},
    doi     = {10.1162/tacl_a_00638}
}

@inproceedings{hsieh2024ruler,
    author    = {Hsieh, Cheng-Ping and Sun, Simeng and Kriman, Samuel and Acharya, Shantanu and Rekesh, Dima and
                 Jia, Fei and Zhang, Yang and Ginsburg, Boris},
    title     = {{RULER}: What's the Real Context Size of Your Long-Context Language Models?},
    booktitle = {First Conference on Language Modeling (COLM)},
    year      = {2024}
}

@inproceedings{wang2025unable,
    author    = {Wang, Chupei and Sun, Jiaqiu Vince},
    title     = {Unable to Forget: Proactive Interference Reveals Working Memory Limits in {LLMs} Beyond Context
                 Length},
    booktitle = {ICML 2025 Workshop on Long Context Foundation Models},
    year      = {2025},
    note      = {arXiv:2506.08184}
}

@inproceedings{wei2022chain,
    author    = {Wei, Jason and Wang, Xuezhi and Schuurmans, Dale and Bosma, Maarten and Ichter, Brian and Xia, Fei and
                 Chi, Ed and Le, Quoc V. and Zhou, Denny},
    title     = {Chain-of-Thought Prompting Elicits Reasoning in Large Language Models},
    booktitle = {Advances in Neural Information Processing Systems},
    volume    = {35},
    year      = {2022}
}

@inproceedings{feng2023towards,
    author    = {Feng, Guhao and Zhang, Bohang and Gu, Yuntian and Ye, Haotian and He, Di and Wang, Liwei},
    title     = {Towards Revealing the Mystery behind Chain of Thought: A Theoretical Perspective},
    booktitle = {Advances in Neural Information Processing Systems},
    volume    = {36},
    year      = {2023}
}

@inproceedings{merrill2024expressive,
    author    = {Merrill, William and Sabharwal, Ashish},
    title     = {The Expressive Power of Transformers with Chain of Thought},
    booktitle = {The Twelfth International Conference on Learning Representations (ICLR)},
    year      = {2024}
}

@inproceedings{kim2023entity,
    author    = {Kim, Najoung and Schuster, Sebastian},
    title     = {Entity Tracking in Language Models},
    booktitle = {Proceedings of the 61st Annual Meeting of the Association for Computational Linguistics (Volume 1:
                 Long Papers)},
    pages     = {3835--3855},
    year      = {2023},
    doi       = {10.18653/v1/2023.acl-long.213}
}

@inproceedings{zheng2024large,
    author    = {Zheng, Chujie and Zhou, Hao and Meng, Fandong and Zhou, Jie and Huang, Minlie},
    title     = {Large Language Models Are Not Robust Multiple Choice Selectors},
    booktitle = {The Twelfth International Conference on Learning Representations (ICLR)},
    year      = {2024}
}

@inproceedings{xie2024adaptive,
    author    = {Xie, Jian and Zhang, Kai and Chen, Jiangjie and Lou, Renze and Su, Yu},
    title     = {Adaptive Chameleon or Stubborn Sloth: Revealing the Behavior of Large Language Models in Knowledge
                 Conflicts},
    booktitle = {The Twelfth International Conference on Learning Representations (ICLR)},
    year      = {2024}
}

@article{kadavath2022language,
    author  = {Kadavath, Saurav and Conerly, Tom and Askell, Amanda and Henighan, Tom and Drain, Dawn and
               Perez, Ethan and Schiefer, Nicholas and Hatfield-Dodds, Zac and DasSarma, Nova and
               Tran-Johnson, Eli and Johnston, Scott and El-Showk, Sheer and Jones, Andy and Elhage, Nelson and
               Hume, Tristan and Chen, Anna and Bai, Yuntao and Bowman, Sam and Fort, Stanislav and
               Ganguli, Deep and Hernandez, Danny and Jacobson, Josh and Kernion, Jackson and Kravec, Shauna and
               Lovitt, Liane and Ndousse, Kamal and Olsson, Catherine and Ringer, Sam and Amodei, Dario and
               Brown, Tom and Clark, Jack and Joseph, Nicholas and Mann, Ben and McCandlish, Sam and
               Olah, Chris and Kaplan, Jared},
    title   = {Language Models (Mostly) Know What They Know},
    journal = {arXiv preprint arXiv:2207.05221},
    year    = {2022}
}

@inproceedings{geifman2017selective,
    author    = {Geifman, Yonatan and El-Yaniv, Ran},
    title     = {Selective Classification for Deep Neural Networks},
    booktitle = {Advances in Neural Information Processing Systems},
    volume    = {30},
    pages     = {4878--4887},
    year      = {2017}
}

@inproceedings{mozannar2020consistent,
    author    = {Mozannar, Hussein and Sontag, David},
    title     = {Consistent Estimators for Learning to Defer to an Expert},
    booktitle = {Proceedings of the 37th International Conference on Machine Learning (ICML)},
    series    = {Proceedings of Machine Learning Research},
    volume    = {119},
    pages     = {7076--7087},
    year      = {2020}
}

@inproceedings{zhang2024rtuning,
    author    = {Zhang, Hanning and Diao, Shizhe and Lin, Yong and Fung, Yi and Lian, Qing and Wang, Xingyao and
                 Chen, Yangyi and Ji, Heng and Zhang, Tong},
    title     = {{R-Tuning}: Instructing Large Language Models to Say `{I} Don't Know'},
    booktitle = {Proceedings of the 2024 Conference of the North American Chapter of the Association for
                 Computational Linguistics: Human Language Technologies (Volume 1: Long Papers)},
    pages     = {7113--7139},
    year      = {2024},
    doi       = {10.18653/v1/2024.naacl-long.394}
}

@inproceedings{yao2023react,
    author    = {Yao, Shunyu and Zhao, Jeffrey and Yu, Dian and Du, Nan and Shafran, Izhak and Narasimhan, Karthik and
                 Cao, Yuan},
    title     = {{ReAct}: Synergizing Reasoning and Acting in Language Models},
    booktitle = {The Eleventh International Conference on Learning Representations (ICLR)},
    year      = {2023}
}

@inproceedings{shi2023large,
    author    = {Shi, Freda and Chen, Xinyun and Misra, Kanishka and Scales, Nathan and Dohan, David and Chi, Ed H. and
                 Sch{\"a}rli, Nathanael and Zhou, Denny},
    title     = {Large Language Models Can Be Easily Distracted by Irrelevant Context},
    booktitle = {Proceedings of the 40th International Conference on Machine Learning (ICML)},
    series    = {Proceedings of Machine Learning Research},
    volume    = {202},
    pages     = {31210--31227},
    year      = {2023}
}

@inproceedings{kuratov2024babilong,
    author    = {Kuratov, Yuri and Bulatov, Aydar and Anokhin, Petr and Rodkin, Ivan and Sorokin, Dmitry and
                 Sorokin, Artyom and Burtsev, Mikhail},
    title     = {{BABILong}: Testing the Limits of {LLMs} with Long Context Reasoning-in-a-Haystack},
    booktitle = {Advances in Neural Information Processing Systems},
    volume    = {37},
    year      = {2024}
}

@inproceedings{merrill2024illusion,
    author    = {Merrill, William and Petty, Jackson and Sabharwal, Ashish},
    title     = {The Illusion of State in State-Space Models},
    booktitle = {Proceedings of the 41st International Conference on Machine Learning (ICML)},
    series    = {Proceedings of Machine Learning Research},
    volume    = {235},
    pages     = {35492--35506},
    year      = {2024}
}

@inproceedings{yang2024latent,
    author    = {Yang, Sohee and Gribovskaya, Elena and Kassner, Nora and Geva, Mor and Riedel, Sebastian},
    title     = {Do Large Language Models Latently Perform Multi-Hop Reasoning?},
    booktitle = {Proceedings of the 62nd Annual Meeting of the Association for Computational Linguistics (Volume 1:
                 Long Papers)},
    pages     = {10210--10229},
    year      = {2024}
}

@inproceedings{zhao2021calibrate,
    author    = {Zhao, Zihao and Wallace, Eric and Feng, Shi and Klein, Dan and Singh, Sameer},
    title     = {Calibrate Before Use: Improving Few-shot Performance of Language Models},
    booktitle = {Proceedings of the 38th International Conference on Machine Learning (ICML)},
    series    = {Proceedings of Machine Learning Research},
    volume    = {139},
    pages     = {12697--12706},
    year      = {2021}
}
